# Least Generalizations and Greatest Specializations of Sets of Clauses

**Shan-Hwei Nienhuys-Cheng** CHENG@CS.FEW.EUR.NL
**Ronald de Wolf** BIDEWOLF@CS.FEW.EUR.NL
*Erasmus University of Rotterdam*
*Department of Computer Science, H4-19*
*P.O. Box 1738, 3000 DR Rotterdam, the Netherlands*

## Abstract

The main operations in Inductive Logic Programming (ILP) are generalization and specialization, which only make sense in a generality order. In ILP, the three most important generality orders are subsumption, implication and implication relative to background knowledge. The two languages used most often are languages of clauses and languages of only Horn clauses. This gives a total of six different ordered languages. In this paper, we give a systematic treatment of the existence or non-existence of least generalizations and greatest specializations of finite sets of clauses in each of these six ordered sets. We survey results already obtained by others and also contribute some answers of our own.

Our main new results are, firstly, the existence of a computable least generalization under implication of every finite set of clauses containing at least one non-tautologous function-free clause (among other, not necessarily function-free clauses). Secondly, we show that such a least generalization need not exist under relative implication, not even if both the set that is to be generalized and the background knowledge are function-free. Thirdly, we give a complete discussion of existence and non-existence of greatest specializations in each of the six ordered languages.

## 1. Introduction

Inductive Logic Programming (ILP) is a subfield of Logic Programming and Machine Learning that tries to induce clausal theories from given sets of positive and negative examples. An inductively inferred theory should imply all of the positive and none of the negative examples. For instance, suppose we are given $P(0)$, $P(s^2(0))$, $P(s^4(0))$, $P(s^6(0))$ as positive examples and $P(s(0)), P(s^3(0)), P(s^5(0))$ as negative examples.[1] Then the set $\Sigma = \{P(0), (P(s^2(x)) \leftarrow P(x))\}$ is a solution: it implies all positive and no negative examples. Note that this set can be seen as a description of the even integers, learned from these examples. Thus induction of clausal theories is a form of learning from examples. For a more extensive introduction to ILP, we refer to (Lavrač & Džeroski, 1994; Muggleton & De Raedt, 1994).

Learning from examples means modifying a theory to bring it more in accordance with the examples. The two main operations in ILP for modification of a theory are *generalization* and *specialization*. Generalization strengthens a theory that is too weak, while specialization weakens a theory that is too strong. These operations only make sense within a *generality order*. This is a relation stating when some clause is more general than some other clause.

1. Here $s^2(0)$ abbreviates $s(s(0))$, $s^3(0)$ abbreviates $s(s(s(0)))$, etc.

The three most important generality orders used in ILP are subsumption (also called $\theta$-subsumption), logical implication and implication relative to background knowledge.[2] In the subsumption order, we say that clause $C$ is more general than $D$—or, equivalently, $D$ is more specific than $C$—in case $C$ subsumes $D$. In the implication order $C$ is more general than $D$ if $C$ logically implies $D$. Finally, $C$ is more general than $D$ relative to background knowledge $\Sigma$ ($\Sigma$ is a set of clauses), if $\{C\} \cup \Sigma$ logically implies $D$.

Of these three orders, subsumption is the most tractable. In particular, subsumption is decidable, whereas logical implication is not decidable, not even for Horn clauses, as established by Marcinkowski and Pacholski (1992). In turn, relative implication is harder than implication: both are undecidable, but proof procedures for implication need to take only derivations from $\{C\}$ into account, whereas a proof procedure for *relative* implication should check all derivations from $\{C\} \cup \Sigma$.

Within a generality order, there are two approaches to generalization or specialization. The first approach generalizes or specializes *individual* clauses. We do not discuss this in any detail in this paper, and merely mention it for completeness' sake. This approach can be traced back to Reynolds' (1970) concept of a *cover*. It was implemented for example by Shapiro (1981) in the subsumption order in the form of *refinement operators*. However, a clause $C$ which implies another clause $D$ need not subsume $D$. For instance, take $C = P(f(x)) \leftarrow P(x)$ and $D = P(f^2(x)) \leftarrow P(x)$. Then $C$ does not subsume $D$, but $C \models D$. Thus subsumption is weaker than implication. A further sign of this weakness is the fact that tautologies need not be subsume-equivalent, even though they are logically equivalent.

The second approach generalizes or specializes *sets* of clauses. This is the approach we will be concerned with in this paper. Here the concept of a *least generalization*[3] is important. The use of such *least* generalizations allows us to generalize cautiously, avoiding over-generalization. Least generalizations of sets of clauses were first discussed by Plotkin (1970, 1971a, 1971b). He proved that any finite set $S$ of clauses has a least generalization under subsumption (LGS). This is a clause which subsumes all clauses in $S$ and which is subsumed by all other clauses that also subsume all clauses in $S$. Positive examples can be generalized by taking their LGS.[4] Of course, we need not take the LGS of *all* positive examples, which would yield a theory consisting of only one clause. Instead, we might divide the positive examples into subsets, and take a separate LGS of each subset. That way we obtain a theory containing more than one clause.

For this second approach, subsumption is again not fully satisfactory. For example, if $S$ consists of the clauses $D_1 = P(f^2(a)) \leftarrow P(a)$ and $D_2 = P(f(b)) \leftarrow P(b)$, then the LGS of $S$ is $P(f(y)) \leftarrow P(x)$. The clause $P(f(x)) \leftarrow P(x)$, which seems more appropriate as a least generalization of $S$, cannot be found by Plotkin's approach, because it does not subsume $D_1$. As this example also shows, the subsumption order is particularly unsatisfactory when we consider *recursive* clauses: clauses which can be resolved with themselves.

2. There is also relative subsumption (Plotkin, 1971b), which will be briefly touched in Section 4.
3. Least generalizations are also often called least *general* generalizations, for instance by Plotkin (1971b), Muggleton and Page (1994), Idestam-Almquist (1993, 1995), Niblett (1988), though not by Plotkin (1970), but we feel this 'general' is redundant.
4. There is also a relation between least generalization under subsumption and *inverse resolution* (Muggleton, 1992).

Because of the weakness of subsumption, it is desirable to make the step from the subsumption order to the more powerful implication order. Accordingly, it is important to find out whether Plotkin's positive result on the existence of LGS's holds for implication as well. So far, the question whether any finite set of clauses has a least generalization under implication (LGI) has only been partly answered. For instance, Idestam-Almquist (1993, 1995) studies least generalizations under *T-implication* as an approximation to LGI's. Muggleton and Page (1994) investigate *self-saturated* clauses. A clause is self-saturated if it is subsumed by any clause which implies it. A clause $D$ is a self-saturation of $C$ if $C$ and $D$ are logically equivalent and $D$ is self-saturated. As Muggleton and Page (1994) state, if two clauses $C_1$ and $C_2$ have self-saturations $D_1$ and $D_2$, then an LGS of $D_1$ and $D_2$ is also an LGI of $C_1$ and $C_2$. This positively answers our question concerning the existence of LGI's in the case of clauses which have a self-saturation. However, Muggleton and Page also show that there exist clauses which have no self-saturation. Hence the concept of self-saturation cannot solve our question in general.

Use of the third generality order, relative implication, is even more desirable than the use of "plain" implication. Relative implication allows us to take background knowledge into account, which can be used to formalize many useful properties and relations of the domain of application. For this reason, least generalizations under implication relative to background knowledge also deserve attention.

Apart from the least generalization, there is also its dual: the *greatest specialization*. Greatest specializations have been accorded much less attention in ILP than least generalizations, but the concept of a greatest specialization may nevertheless be useful (see the beginning of Section 6).

In this paper, we give a systematic treatment of the existence and non-existence of least generalizations and greatest specializations, applied to each of these three generality orders. Apart from distinguishing between these three orders, we also distinguish between languages of general clauses and more restricted languages of Horn clauses. Though most researchers in ILP restrict attention to Horn clauses, general clauses are also sometimes used (Plotkin, 1970, 1971b; Shapiro, 1981; De Raedt & Bruynooghe, 1993; Idestam-Almquist, 1993, 1995). Moreover, many researchers who do not use general clauses actually allow negative literals to appear in the body of a clause. That is, they use clauses of the form $A \leftarrow L_1, \ldots, L_n$, where $A$ is an atom and each $L_i$ is a literal. These are called *program* clauses (Lloyd, 1987). Program clauses are in fact logically equivalent to general clauses. For instance, the program clause $P(x) \leftarrow Q(x), \neg R(x)$ is equivalent to the non-Horn clause $P(x) \vee \neg Q(x) \vee R(x)$. For these two reasons we consider not only languages of Horn clauses, but also pay attention to languages of general clauses.

The combination of three generality orders and two different possible languages of clauses gives a total of six different ordered languages. For each of these, we can ask whether least generalizations (LG's) and greatest specializations (GS's) always exist. We survey results already obtained by others and also contribute some answers of our own. For the sake of clarity, we will summarize the results of our survey right at the outset. In the following table '+' signifies a positive answer, and '−' means a negative answer.

| Quasi-order | Horn clauses | | General clauses | |
|---|---|---|---|---|
| | LG | GS | LG | GS |
| Subsumption ($\succeq$) | + | + | + | + |
| Implication ($\models$) | − | − | + for function-free | + |
| Relative implication ($\models_\Sigma$) | − | − | − | + |

Table 1: Existence of LG's and GS's

Our own contributions to this table are threefold. First and foremost, we prove that if $S$ is a finite set of clauses containing at least one non-tautologous function-free clause[5] (apart from this non-tautologous function-free clause, $S$ may contain an arbitrary finite number of other clauses, including clauses which contain functions), then there exists a computable LGI of $S$. This result is on the one hand based on the Subsumption Theorem for resolution (Lee, 1967; Kowalski, 1970; Nienhuys-Cheng & de Wolf, 1996), which allows us to restrict attention to finite sets of ground instances of clauses, and on the other hand on a modification of some proofs concerning T-implication which can be found in (Idestam-Almquist, 1993, 1995). An immediate corollary of this result is the existence and computability of an LGI of any finite set of function-free clauses. As far as we know, both our general LGI-result and this particular corollary are new results.

Niblett (1988, p. 135) claims that "it is simple to show that there are lggs if the language is restricted to a fixed set of constant symbols since all Herbrand interpretations are finite." Yet even for this special case of our general result, it appears that no proof has been published. Initially, we found a direct proof of this case, but this was not really any simpler than the proof of the more general result that we give in this paper. Niblett's idea that the proof is simple may be due to some confusion about the relation between Herbrand models and logical implication (which is defined in terms of *all* models, not just Herbrand models). We will describe this at the end of Subsection 5.1. Or perhaps one might think that the decidability of implication for function-free clauses immediately implies the existence of an LGI. But in fact, decidability is not a sufficient condition for the existence of a least generalization. For example, it is decidable whether one function-free clause $C$ implies another function-free clause $D$ relative to function-free background knowledge. Yet least generalizations relative to function-free background knowledge do not always exist, as we will show in Section 7.

Our LGI-result does not solve the general question of the existence of LGI's, but it does provide a positive answer for a large class of cases: the presence of one non-tautologous function-free clause in a finite $S$ already guarantees the existence and computability of an LGI of $S$, no matter what other clauses $S$ contains.[6] Because of the prominence of function-free clauses in ILP, this case may be of great practical signifcance. Often, particularly in implementations of ILP-systems, the language is required to be function-free, or function

5. A clause which only contains constants and variables as terms.
6. Note that even for function-free clauses, the subsumption order is still not enough. Consider $D_1 = P(x,y,z) \leftarrow P(y,z,x)$ and $D_2 = P(x,y,z) \leftarrow P(z,x,y)$ (this example is adapted from Idestam-Almquist). $D_1$ is a resolvent of $D_2$ and $D_2$ and $D_2$ is a resolvent of $D_1$ and $D_1$. Hence $D_1$ and $D_2$ are logically equivalent. This means that $D_1$ is an LGI of the set $\{D_1, D_2\}$. However, the LGS of these two clauses is $P(x,y,z) \leftarrow P(u,v,w)$, which is clearly an over-generalization.

symbols are removed from clauses and put in the background knowledge by techniques such as *flattening* (Rouveirol, 1992). Well-known ILP-systems such as Foil (Quinlan & Cameron-Jones, 1993), Linus (Lavrač & Džeroski, 1994) and Mobal (Morik, Wrobel, Kietz, & Emde, 1993) all use only function-free clauses. More than one half of the ILP-systems surveyed by Aha (1992) is restricted to function-free clauses. Function-free clauses are also sufficient for most applications concerning databases.

Our second contribution shows that a set $S$ need not have a least generalization relative to some background knowledge $\Sigma$, not even when $S$ and $\Sigma$ are both function-free.

Thirdly, we contribute a complete discussion of existence and non-existence of greatest specializations in each of the six ordered languages. In particular, we show that any finite set of clauses has a greatest specialization under implication. Combining this with the corollary of our result on LGI's, it follows that a function-free clausal language is a *lattice*.

## 2. Preliminaries

In this section we will define some of the concepts we need. For the definitions of 'model', 'tautology', 'substitution', etc., we refer to standard works such as (Chang & Lee, 1973; Lloyd, 1987). A *positive literal* is an atom, a *negative literal* is the negation of an atom. A *clause* is a finite set of literals, which is treated as the universally quantified disjunction of those literals. A *definite program clause* is a clause with one positive and zero or more negative literals and a *definite goal* is a clause without positive literals. A *Horn clause* is either a definite program clause or a definite goal. If $C$ is a clause, we use $C^+$ to denote the positive literals in $C$, and $C^-$ to denote the negative literals in $C$. The empty clause, which represents a contradiction, is denoted by $\square$.

**Definition 1** Let $\mathcal{A}$ be an alphabet of the first-order logic. Then the *clausal language* $\mathcal{C}$ *by* $\mathcal{A}$ is the set of all clauses which can be constructed from the symbols in $\mathcal{A}$. The *Horn language* $\mathcal{H}$ *by* $\mathcal{A}$ is the set of all Horn clauses which can be constructed from the symbols in $\mathcal{A}$. $\square$

In this paper, we just presuppose some arbitrary alphabet $\mathcal{A}$, and consider the clausal language $\mathcal{C}$ and Horn language $\mathcal{H}$ based on this $\mathcal{A}$. We will now define three increasingly strong generality orders on clauses: subsumption, implication and relative implication.

**Definition 2** Let $C$ and $D$ be clauses and $\Sigma$ be a set of clauses. We say that $C$ *subsumes* $D$, denoted as $C \succeq D$, if there exists a substitution $\theta$ such that $C\theta \subseteq D$.[7] $C$ and $D$ are *subsume-equivalent* if $C \succeq D$ and $D \succeq C$.

$\Sigma$ *(logically) implies* $C$, denoted as $\Sigma \models C$, if every model of $\Sigma$ is also a model of $C$. $C$ *(logically) implies* $D$, denoted as $C \models D$, if $\{C\} \models D$. $C$ and $D$ are *(logically) equivalent* if $C \models D$ and $D \models C$.

$C$ *implies* $D$ *relative to* $\Sigma$, denoted as $C \models_\Sigma D$, if $\Sigma \cup \{C\} \models D$. $C$ and $D$ are *equivalent relative to* $\Sigma$ if $C \models_\Sigma D$ and $D \models_\Sigma C$. $\square$

7. Right from the very first applications of subsumption in ILP, there has been some controversy about the symbol used for subsumption: Plotkin (1970) used '$\leq$', while Reynolds (1970) used '$\geq$'. We use '$\succeq$' here, similar to Reynolds' '$\geq$', because we feel it serves the intuition to view $C$ as somehow "bigger" or "stronger" than $D$, if $C \succeq D$ holds.

If $C$ does not subsume $D$, we write $C \not\succeq D$. Similarly, we use the notation $C \not\models D$ and $C \not\models_\Sigma D$.

If $C \succeq D$, then $C \models D$. The converse does not hold, as the examples in the introduction showed. Similarly, if $C \models D$, then $C \models_\Sigma D$, and again the converse need not hold. Consider $C = P(a) \vee \neg P(b)$, $D = P(a)$, and $\Sigma = \{P(b)\}$: then $C \models_\Sigma D$, but $C \not\models D$.

We now proceed to define a proof procedure for logical implication between clauses, using resolution and subsumption.

**Definition 3** If two clauses have no variables in common, then they are said to be *standardized apart.*

Let $C_1 = L_1 \vee \ldots \vee L_i \vee \ldots \vee L_m$ and $C_2 = M_1 \vee \ldots \vee M_j \vee \ldots \vee M_n$ be two clauses which are standardized apart. If the substitution $\theta$ is a most general unifier (mgu) of the set $\{L_i, \neg M_j\}$, then the clause $((C_1 - L_i) \cup (C_2 - M_j))\theta$ is called a *binary resolvent* of $C_1$ and $C_2$. The literals $L_i$ and $M_j$ are said to be the literals *resolved upon.* □

If $C_1$ and $C_2$ are not standardized apart, we can take a variant $C_2'$ of $C_2$, such that $C_1$ and $C_2'$ are standardized apart. For simplicity, a binary resolvent of $C_1$ and $C_2'$ is also called a binary resolvent of $C_1$ and $C_2$ itself.

**Definition 4** Let $C$ be a clause and $\theta$ an mgu of $\{L_1, \ldots, L_n\} \subseteq C$ $(n \geq 1)$. Then the clause $C\theta$ is called a *factor* of $C$. □

Note that any non-empty clause $C$ is a factor of itself, using the empty substitution $\varepsilon$ as an mgu of a single literal in $C$.

**Definition 5** A *resolvent* $C$ of clauses $C_1$ and $C_2$ is a binary resolvent of a factor of $C_1$ and a factor of $C_2$, where the literals resolved upon are the literals unified in the respective factors. $C_1$ and $C_2$ are the *parent clauses* of $C$. □

**Definition 6** Let $\Sigma$ be a set of clauses and $C$ a clause. A *derivation* of $C$ from $\Sigma$ is a finite sequence of clauses $R_1, \ldots, R_k = C$, such that each $R_i$ is either in $\Sigma$, or a resolvent of two clauses in $\{R_1, \ldots, R_{i-1}\}$. If such a derivation exists, we write $\Sigma \vdash_r C$. □

**Definition 7** Let $\Sigma$ be a set of clauses and $C$ a clause. We say there exists a *deduction* of $C$ from $\Sigma$, written as $\Sigma \vdash_d C$, if $C$ is a tautology, or if there exists a clause $D$ such that $\Sigma \vdash_r D$ and $D \succeq C$. □

The next result, proved by Nienhuys-Cheng and de Wolf (1996), generalizes Herbrand's Theorem:

**Theorem 1** *Let $\Sigma$ be a set of clauses and $C$ be a ground clause. If $\Sigma \models C$, then there exists a finite set $\Sigma_g$ of ground instances of clauses in $\Sigma$, such that $\Sigma_g \models C$.*

The following Subsumption Theorem gives a precise characterization of implication between clauses in terms of resolution and subsumption. It was proved by Lee (1967), Kowalski (1970) and reproved by Nienhuys-Cheng and de Wolf (1996).

**Theorem 2 (Subsumption theorem)** *Let $\Sigma$ be a set of clauses and $C$ be a clause. Then $\Sigma \models C$ iff $\Sigma \vdash_d C$.*

The next lemma was first proved by Gottlob (1987). Actually, it is an immediate corollary of the subsumption theorem:

**Lemma 1 (Gottlob)** *Let $C$ and $D$ be non-tautologous clauses. If $C \models D$, then $C^+ \succeq D^+$ and $C^- \succeq D^-$.*

**Proof** Since $C^+ \succeq C$, if $C \models D$, then we have $C^+ \models D$. Since $C^+$ cannot be resolved with itself, it follows from the subsumption theorem that $C^+ \succeq D$. But then $C^+$ must subsume the positive literals in $D$, hence $C^+ \succeq D^+$. Similarly $C^- \succeq D^-$. □

An important consequence of this lemma concerns the *depth* of clauses, defined as follows:

**Definition 8** Let $t$ be a term. If $t$ is a variable or constant, then the *depth* of $t$ is 1. If $t = f(t_1, \ldots, t_n)$, $n \geq 1$, then the depth of $t$ is 1 plus the depth of the $t_i$ with largest depth. The *depth* of a clause $C$ is the depth of the term with largest depth in $C$. □

For example, the term $t = f(a, x)$ has depth 2. $C = P(f(x)) \leftarrow P(g(f(x), a))$ has depth 3, since $g(f(x), a)$ has depth 3. It follows from Gottlob's lemma that if $C \models D$, then the depth of $C$ is smaller than or equal to the depth of $D$, for otherwise $C^+$ cannot subsume $D^+$ or $C^-$ cannot subsume $D^-$. For instance, take $D = P(x, f(x, g(y))) \leftarrow P(g(a), b)$, which has depth 3. Then a clause $C$ containing a term $f(x, g^2(y))$ (depth 4) cannot imply $D$.

**Definition 9** Let $S$ and $S'$ be finite sets of clauses, $x_1, \ldots, x_n$ all distinct variables appearing in $S$, and $a_1, \ldots, a_n$ distinct constants not appearing in $S$ or $S'$. Then $\sigma = \{x_1/a_1, \ldots, x_n/a_n\}$ is called a *Skolem substitution* for $S$ w.r.t. $S'$. If $S'$ is empty, we just say that $\sigma$ is a Skolem substitution for $S$. □

**Lemma 2** *Let $\Sigma$ be a set of clauses, $C$ be a clause, and $\sigma$ be a Skolem substitution for $C$ w.r.t. $\Sigma$. Then $\Sigma \models C$ iff $\Sigma \models C\sigma$.*

**Proof**

$\Rightarrow$: Obvious.

$\Leftarrow$: Suppose $C$ is not a tautology and let $\sigma = \{x_1/a_1, \ldots, x_n/a_n\}$. If $\Sigma \models C\sigma$, it follows from the subsumption theorem that there is a $D$ such that $\Sigma \vdash_r D$ and $D \succeq C\sigma$. Thus there is a $\theta$, such that $D\theta \subseteq C\sigma$. Note that since $\Sigma \vdash_r D$ and none of the constants $a_1, \ldots, a_n$ appears in $\Sigma$, none of these constants appears in $D$. Now let $\theta'$ be obtained by replacing in $\theta$ all occurrences of $a_i$ by $x_i$, for every $1 \leq i \leq n$. Then $D\theta' \subseteq C$, hence $D \succeq C$. Therefore $\Sigma \vdash_d C$ and hence $\Sigma \models C$. □

## 3. Least Generalizations and Greatest Specializations

In this section, we will define the concepts we need concerning least generalizations and greatest specializations.

**Definition 10** Let $\Gamma$ be a set and $R$ be a binary relation on $\Gamma$.

1. $R$ is *reflexive on* $\Gamma$, if $xRx$ for every $x \in \Gamma$.
2. $R$ is *transitive on* $\Gamma$, if for every $x, y, z \in \Gamma$, $xRy$ and $yRz$ implies $xRz$.
3. $R$ is *symmetric on* $\Gamma$, if for every $x, y \in \Gamma$, $xRy$ implies $yRx$.
4. $R$ is *anti-symmetric on* $\Gamma$, if for every $x, y, z \in \Gamma$, $xRy$ and $yRx$ implies $x = y$.

If $R$ is both reflexive and transitive on $\Gamma$, we say $R$ is a *quasi-order* on $\Gamma$. If $R$ is both reflexive, transitive and anti-symmetric on $\Gamma$, we say $R$ is a *partial order* on $\Gamma$. If $R$ is reflexive, transitive and symmetric on $\Gamma$, $R$ is an *equivalence relation* on $\Gamma$. □

A quasi-order $R$ on $\Gamma$ induces an equivalence-relation $\sim$ on $\Gamma$, as follows: we say $x, y \in \Gamma$ are *equivalent* induced by $R$ (denoted $x \sim y$) if both $xRy$ and $yRx$. Using this equivalence relation, a quasi-order $R$ on $\Gamma$ induces a partial order $R'$ on the set of equivalence classes in $\Gamma$, defined as follows: if $[x]$ denotes the equivalence class of $x$ (i.e., $[x] = \{y \mid x \sim y\}$), then $[x]R'[y]$ iff $xRy$.

We first give a general definition of least generalizations and greatest specializations for sets of clauses ordered by some quasi-order, which we then instantiate in different ways.

**Definition 11** Let $\Gamma$ be a set of clauses, $\geq$ be a quasi-order on $\Gamma$, $S \subseteq \Gamma$ be a finite set of clauses and $C \in \Gamma$. If $C \geq D$ for every $D \in S$, then we say $C$ is a *generalization* of $S$ under $\geq$. Such a $C$ is called a *least generalization (LG)* of $S$ under $\geq$ in $\Gamma$, if we have $C' \geq C$ for every generalization $C' \in \Gamma$ of $S$ under $\geq$.

Dually, $C$ is a *specialization* of $S$ under $\geq$, if $D \geq C$ for every $D \in S$. Such a $C$ is called a *greatest specialization (GS)* of $S$ under $\geq$ in $\Gamma$, if we have $C \geq C'$ for every specialization $C' \in \Gamma$ of $S$ under $\geq$. □

It is easy to see that if some set $S$ has an LG or GS under $\geq$ in $\Gamma$, then this LG or GS will be unique up to the equivalence induced by $\geq$ in $\Gamma$. That is, if $C$ and $D$ are both LG's or GS's of some set $S$, then we have $C \sim D$.

The concepts defined above are instances of the mathematical concepts of (least) upper bounds and (greatest) lower bounds. Thus we can speak of lattice-properties of a quasi- or partially ordered set of clauses:

**Definition 12** Let $\Gamma$ be a set of clauses and $\geq$ be a quasi-order on $\Gamma$. If for every finite subset $S$ of $\Gamma$, there exist both a least generalization and a greatest specialization of $S$ under $\geq$ in $\Gamma$, then the set $\Gamma$ ordered by $\geq$ is called a *lattice*. □

It should be noted that usually in mathematics, a lattice is defined for a partial order instead of a quasi-order. However, since in ILP we usually have to deal with individual clauses rather than with equivalence classes of clauses, it is convenient for us to define 'lattice' for a quasi-order here. Anyhow, if a quasi-order $\geq$ is a lattice on $\Gamma$, then the partial order induced by $\geq$ is a lattice on the set of equivalence classes in $\Gamma$.

In ILP, there are two main instantiations for the set of clauses $\Gamma$: either we take a clausal language $\mathcal{C}$, or we take a Horn language $\mathcal{H}$. Similarly, there are three interesting choices for the quasi-order $\geq$: we can use either $\succeq$ (subsumption), $\models$ (implication), or $\models_\Sigma$ (relative implication) for some background knowledge $\Sigma$. In the $\succeq$-order, we will sometimes abbreviate the terms 'least generalization of $S$ under subsumption' and 'greatest specialization of $S$ under subsumption' to 'LGS of $S$' and 'GSS of $S$', respectively. Similarly, in the $\models$-order we will sometimes speak of an LGI (least generalization under implication) and a GSI. In the $\models_\Sigma$-order, we will use LGR (least generalization under *relative* implication) and GSR.

These two different languages and three different quasi-orders give a total of six combinations. For each combination, we can ask whether an LG or GS of every finite set $S$ exists. In the next section, we will review the answers for subsumption given by others or by ourselves. Then we devote two sections to least generalizations and greatest specializations under implication, respectively. Finally, we discuss least generalizations and greatest specializations under relative implication. The results of this survey have already been summarized in Table 1 in the introduction.

## 4. Subsumption

First we devote some attention to subsumption. Least generalizations under subsumption have been discussed extensively by Plotkin (1970). The main result in Plotkin's framework is the following:

**Theorem 3 (Existence of LGS in $\mathcal{C}$)** *Let $\mathcal{C}$ be a clausal language. Then for every finite $S \subseteq \mathcal{C}$, there exists an LGS of $S$ in $\mathcal{C}$.*

If $S$ only contains Horn clauses, then it can be shown that the LGS of $S$ is itself also a Horn clause. Thus the question for the existence of an LGS of every finite set $S$ of clauses is answered positively for both clausal languages and for Horn languages.

Plotkin established the existence of an LGS, but he seems to have ignored the GSS in (1970, 1971b), possibly because it is a very straightforward result. It is in fact fairly easy to show that the GSS of some finite set $S$ of clauses is simply the union of all clauses in $S$ after they are standardized apart.[8] We include the proof here.

**Theorem 4 (Existence of GSS in $\mathcal{C}$)** *Let $\mathcal{C}$ be a clausal language. Then for every finite $S \subseteq \mathcal{C}$, there exists a GSS of $S$ in $\mathcal{C}$.*

**Proof** Suppose $S = \{D_1, \ldots, D_n\} \subseteq \mathcal{C}$. Without loss of generality, we assume the clauses in $S$ are standardized apart. Let $D = D_1 \cup \ldots \cup D_n$, then $D_i \succeq D$, for every $1 \leq i \leq n$. Now let $C \in \mathcal{C}$ be such that $D_i \succeq C$, for every $1 \leq i \leq n$. Then for every $1 \leq i \leq n$, there is a $\theta_i$ such that $D_i\theta_i \subseteq C$ and $\theta_i$ only acts on variables in $D_i$. If we let $\theta = \theta_1 \cup \ldots \cup \theta_n$, then $D\theta = D_1\theta_1 \cup \ldots \cup D_n\theta_n \subseteq C$. Hence $D \succeq C$, so $D$ is a GSS of $S$ in $\mathcal{C}$. □

8. Note that this has nothing to do with *unification*. For instance, if $S = \{P(a,x), P(y,b)\}$, then the GSS of $S$ in $\mathcal{C}$ would be $P(a,x) \vee P(y,b)$. However, if we would instantiate $\Gamma$ in Definition 11 to the set of atoms, then the greatest specialization of two atoms *in the set of atoms* should itself also be an atom. The GSS of two atoms is then their most general unification (Reynolds, 1970). For instance, the GSS of $S$ would in this case be $P(a,b)$.

This establishes that a clausal language $\mathcal{C}$ ordered by $\succeq$ is a lattice.

Proving the existence of a GSS of every finite set of Horn clauses in $\mathcal{H}$ requires a little more work, but here also the result is positive. For example, $D = P(a) \leftarrow P(f(a)), Q(y)$ is a GSS of $D_1 = P(x) \leftarrow P(f(x))$ and $D_2 = P(a) \leftarrow Q(y)$. Note that $D$ can be obtained by applying $\sigma = \{x/a\}$ (the mgu of the heads of $D_1$ and $D_2$) to $D_1 \cup D_2$, the GSS of $D_1$ and $D_2$ in $\mathcal{C}$. This idea will be used in the following proof. Here we assume $\mathcal{H}$ contains an artificial bottom element (True) $\perp$, such that $C \succeq \perp$ for every $C \in \mathcal{H}$, and $\perp \not\succeq C$ for every $C \neq \perp$. Note that $\perp$ is not subsume-equivalent with other tautologies.

**Theorem 5 (Existence of GSS in $\mathcal{H}$)** *Let $\mathcal{H}$ be a Horn language, with $\perp \in \mathcal{H}$. Then for every finite $S \subseteq \mathcal{H}$, there exists a GSS of $S$ in $\mathcal{H}$.*

**Proof** Suppose $S = \{D_1, \ldots, D_n\} \subseteq \mathcal{H}$. Without loss of generality we assume the clauses in $S$ are standardized apart, $D_1, \ldots, D_k$ are the definite program clauses in $S$, and $D_{k+1}, \ldots, D_n$ are the definite goals in $S$. If $k = 0$ (i.e., if $S$ only contains goals), then it is easy to show that $D_1 \cup \ldots \cup D_n$ is a GSS of $S$ in $\mathcal{H}$. If $k \geq 1$ and the set $\{D_1^+, \ldots, D_k^+\}$ is not unifiable, then $\perp$ is a GSS of $S$ in $\mathcal{H}$. Otherwise, let $\sigma$ be an mgu of $\{D_1^+, \ldots, D_k^+\}$, and let $D = D_1\sigma \cup \ldots \cup D_n\sigma$ (note that actually $D_i\sigma = D_i$ for $k+1 \leq i \leq n$, since the clauses in $S$ are standardized apart). Since $D$ has exactly one literal in its head, it is a definite program clause. Furthermore, we have $D_i \succeq D$ for every $1 \leq i \leq n$, since $D_i\sigma \subseteq D$.

To show that $D$ is a GSS of $S$ in $\mathcal{H}$, suppose $C \in \mathcal{H}$ is some clause such that $D_i \succeq C$ for every $1 \leq i \leq n$. For every $1 \leq i \leq n$, let $\theta_i$ be such that $D_i\theta_i \subseteq C$ and $\theta_i$ only acts on variables in $D_i$. Let $\theta = \theta_1 \cup \ldots \cup \theta_n$. For every $1 \leq i \leq k$, $D_i^+\theta = D_i^+\theta_i = C^+$, so $\theta$ is a unifier of $\{D_1^+, \ldots, D_k^+\}$. But $\sigma$ is an mgu of this set, so there is a $\gamma$ such that $\theta = \sigma\gamma$. Now $D\gamma = D_1\sigma\gamma \cup \ldots \cup D_n\sigma\gamma = D_1\theta \cup \ldots \cup D_n\theta = D_1\theta_1 \cup \ldots \cup D_n\theta_n \subseteq C$. Hence $D \succeq C$, so $D$ is a GSS of $S$ in $\mathcal{H}$. See figure 1 for illustration of the case where $n = 2$.

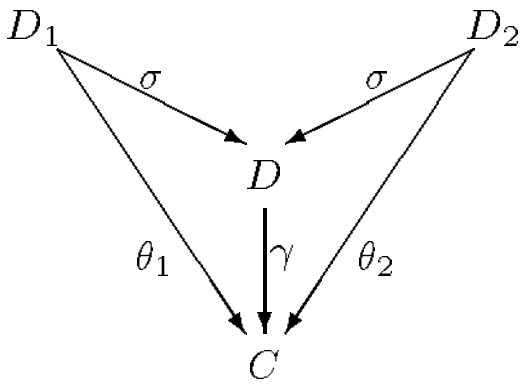

Figure 1: $D$ is a GSS of $D_1$ and $D_2$

□

Thus a Horn language $\mathcal{H}$ ordered by $\succeq$ is also a lattice.

We end this section by briefly discussing Plotkin's (1971b) *relative subsumption.* This is an extension of subsumption which takes background knowledge into account. This background knowledge is rather restricted: it must be a finite set $\Sigma$ of *ground literals.* Because of its restrictiveness, we have not included relative subsumption in Table 1. Nevertheless, we mention it here, because least generalization under relative subsumption forms the basis of the well-known ILP system GOLEM (Muggleton & Feng, 1992).

**Definition 13** Let $C, D$ be clauses, $\Sigma = \{L_1, \ldots, L_m\}$ be a finite set of ground literals. Then *$C$ subsumes $D$ relative to* $\Sigma$, denoted by $C \succeq_\Sigma D$, if $C \succeq (D \cup \{\neg L_1, \ldots, \neg L_m\})$. □

It is easy to see that $\succeq_\Sigma$ is reflexive and transitive, so it imposes a quasi-order on a set of clauses.

Suppose $S = \{D_1, \ldots, D_n\}$ and $\Sigma = \{L_1, \ldots, L_m\}$. It is easy to see that an LGS of $\{(D_1 \cup \{\neg L_1, \ldots, \neg L_m\}), \ldots, (D_n \cup \{\neg L_1, \ldots, \neg L_m\})\}$ is a least generalization of $S$ under $\succeq_\Sigma$, so every finite set of clauses has a least generalization under $\succeq_\Sigma$ in $\mathcal{C}$. Moreover, if each $D_i$ is a Horn clause and each $L_j$ is a *positive* ground literal (i.e., a ground atom), then this least generalization will itself also be a Horn clause. Accordingly, if $\Sigma$ is a finite set of positive ground literals, then every finite set of Horn clauses has a least generalization under $\succeq_\Sigma$ in $\mathcal{H}$.

## 5. Least Generalizations under Implication

Now we turn from subsumption to the implication order. In this section we will discuss LGI's, in the next section we handle GSS's. For Horn clauses, the LGI-question has already been answered negatively by Muggleton and De Raedt (1994).

Let $D_1 = P(f^2(x)) \leftarrow P(x)$, $D_2 = P(f^3(x)) \leftarrow P(x)$, $C_1 = P(f(x)) \leftarrow P(x)$ and $C_2 = P(f^2(y)) \leftarrow P(x)$. Then we have both $C_1 \models \{D_1, D_2\}$ and $C_2 \models \{D_1, D_2\}$. It is not very difficult to see that there are no more specific Horn clauses than $C_1$ and $C_2$ that imply both $D_1$ and $D_2$. For $C_1$: no resolvent of $C_1$ with itself implies $D_2$ and no clause that is properly subsumed by $C_1$ still implies $D_1$ and $D_2$. For $C_2$: every resolvent of $C_2$ with itself is a variant of $C_2$, and no clause that is properly subsumed by $C_2$ still implies $D_1$ and $D_2$. Thus $C_1$ and $C_2$ are both "minimal" generalizations under implication of $\{D_1, D_2\}$. Since $C_1$ and $C_2$ are not logically equivalent under implication, there is no LGI of $\{D_1, D_2\}$ in $\mathcal{H}$.

However, the fact that there is no LGI of $\{D_1, D_2\}$ in $\mathcal{H}$ does not mean that $D_1$ and $D_2$ have no LGI in $\mathcal{C}$, since a Horn language is a more restricted space than a clausal language. In fact, it is shown by Muggleton and Page (1994) that $C = P(f(x)) \vee P(f^2(y)) \leftarrow P(x)$ is an LGI of $D_1$ and $D_2$ in $\mathcal{C}$. For this reason, it may be worthwhile for the LGI to consider a clausal language instead of only Horn clauses.

In the next subsection, we show that any finite set of clauses which contains at least one non-tautologous function-free clause, has an LGI in $\mathcal{C}$. An immediate corollary of this result is the existence of an LGI of any finite set of function-free clauses. In our usage of the word, a 'function-free' clause may contain constants, even though constants are sometimes seen as functions of arity 0.

**Definition 14** A clause is *function-free* if it does not contain function symbols of arity 1 or more. □

Note that a clause is function-free iff it has depth 1. In case of sets of clauses which all contain function symbols, the LGI-question remains open.

### 5.1 A Sufficient Condition for the Existence of an LGI

In this subsection, we will show that any finite set $S$ of clauses containing at least one non-tautologous function-free clause, has an LGI in $\mathcal{C}$.

**Definition 15** Let $C$ be a clause, $x_1, \ldots, x_n$ all distinct variables in $C$, and $K$ a set of terms. Then the *instance set* of $C$ w.r.t. $K$ is $\mathcal{I}(C, K) = \{C\theta \mid \theta = \{x_1/t_1, \ldots, x_n/t_n\},$ where $t_i \in K$, for every $1 \leq i \leq n\}$. If $\Sigma = \{C_1, \ldots, C_k\}$ is a set of clauses, then the *instance set* of $\Sigma$ w.r.t. $K$ is $\mathcal{I}(\Sigma, K) = \mathcal{I}(C_1, K) \cup \ldots \cup \mathcal{I}(C_k, K)$. □

For example, if $C = P(x) \vee Q(y)$ and $T = \{a, f(z)\}$, then $\mathcal{I}(C, T) = \{(P(a) \vee Q(a)), (P(a) \vee Q(f(z))), (P(f(z)) \vee Q(a)), (P(f(z)) \vee Q(f(z)))\}$.

**Definition 16** Let $S$ be a finite set of clauses, and $\sigma$ be a Skolem substitution for $S$. Then the *term set* of $S$ by $\sigma$ is the set of all terms (including subterms) occurring in $S\sigma$. □

A term set of $S$ by some $\sigma$ is a finite set of ground terms. For instance, the term set of $D = P(f^2(x), y, z) \leftarrow P(y, z, f^2(x))$ by $\sigma = \{x/a, y/b, z/c\}$ is $T = \{a, f(a), f^2(a), b, c\}$.

Our definition of a term set corresponds to what Idestam-Almquist (1993, 1995) calls a 'minimal term set'. In his definition, if $\sigma$ is a Skolem substitution for a set of clauses $S = \{D_1, \ldots, D_n\}$ w.r.t. some other set of clauses $S'$, then a *term set* of $S$ is a finite set of terms which contains the minimal term set of $S$ by $\sigma$ as a subset.

Using his notion of term set, he defines *T-implication* as follows: if $C$ and $D$ are clauses and $T$ is a term set of $\{D\}$ by some Skolem substitution $\sigma$ w.r.t. $\{C\}$, then $C$ *T-implies* $D$ w.r.t. $T$ if $\mathcal{I}(C, T) \models D\sigma$. T-implication is decidable, weaker than logical implication and stronger than subsumption. Idestam-Almquist (1993, 1995) gives the result that any finite set of clauses has a least generalization under T-implication w.r.t. any term set $T$. However, as he also notes, T-implication is not transitive and hence not a quasi-order. Therefore it does not fit into our general framework here. For this reason, we will not discuss it fully here, and for the same reason we have not included a row for T-implication in Table 1.

Let us now begin with the proof of our result concerning the existence of LGI's. Consider $C = P(x, y, z) \leftarrow P(z, x, y)$ and $D$, $\sigma$ and $T$ as above. Then $C \models D$ and also $\mathcal{I}(C, T) \models D\sigma$, since $D\sigma$ is a resolvent of $P(f^2(a), b, c) \leftarrow P(c, f^2(a), b)$ and $P(c, f^2(a), b) \leftarrow P(b, c, f^2(a))$, which are in $\mathcal{I}(C, T)$. As we will show in the next lemma, this holds in general: if $C \models D$ and $C$ is function-free, then we can restrict attention to the ground instances of $C$ instantiated to terms in the term set of $D$ by some $\sigma$.

The proof of Lemma 3 uses the following idea. Consider a derivation of a clause $E$ from a set $\Sigma$ of ground clauses. Suppose some of the clauses in $\Sigma$ contain terms not appearing in $E$. Then any literals containing these terms in $\Sigma$ must be resolved away in the derivation. This means that if we replace all the terms in the derivation that are not in $E$, by some other term $t$, then the result will be another derivation of $E$. For example, the left of figure 2 shows a derivation of length 1 of $E$. The term $f^2(b)$ in the parent clauses does not appear in $E$. If we replace this term by the constant $a$, the result is another derivation of $E$ (right of the figure).

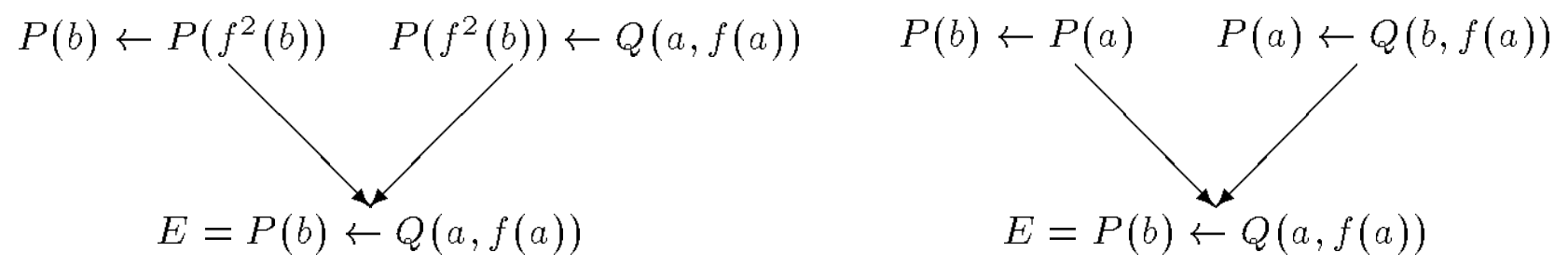

Figure 2: Transforming the left derivation yields the right derivation

**Lemma 3** *Let $C$ be a function-free clause, $D$ be a clause, $\sigma$ be a Skolem substitution for $D$ w.r.t. $\{C\}$ and $T$ be the term set of $D$ by $\sigma$. Then $C \models D$ iff $\mathcal{I}(C, T) \models D\sigma$.*

**Proof**

$\Leftarrow$: Since $C \models \mathcal{I}(C,T)$ and $\mathcal{I}(C,T) \models D\sigma$, we have $C \models D\sigma$. Now $C \models D$ by Lemma 2.

$\Rightarrow$: If $D$ is a tautology, then $D\sigma$ is a tautology, so this case is obvious. Suppose $D$ is not a tautology, then $D\sigma$ is not a tautology. Since $C \models D\sigma$, it follows from Theorem 1 that there exists a finite set $\Sigma$ of ground instances of $C$, such that $\Sigma \models D\sigma$. By the Subsumption Theorem, there exists a derivation from $\Sigma$ of a clause $E$, such that $E \succeq D\sigma$. Since $\Sigma$ is ground, $E$ must also be ground, so we have $E \subseteq D\sigma$. This implies that $E$ only contains terms from $T$.

Let $t$ be an arbitrary term in $T$ and let $\Sigma'$ be obtained from $\Sigma$ by replacing every term in clauses in $\Sigma$ which is not in $T$, by $t$. Note that since each clause in $\Sigma$ is a ground instance of the function-free clause $C$, every clause in $\Sigma'$ is also a ground instance of $C$. Now it is easy to see that the same replacement of terms in the derivation of $E$ from $\Sigma$ results in a derivation of $E$ from $\Sigma'$: (1) each resolution step in the derivation from $\Sigma$ can also be carried out in the derivation from $\Sigma'$, since the same terms in $\Sigma$ are replaced by the same terms in $\Sigma'$, and (2) the terms in $\Sigma$ that are not in $T$ (and hence are replaced by $t$) do not appear in the conclusion $E$ of the derivation.

Since there is a derivation of $E$ from $\Sigma$ we have $\Sigma' \models E$, and hence $\Sigma' \models D\sigma$. $\Sigma'$ is a set of ground instances of $C$ and all terms in $\Sigma'$ are terms in $T$, so $\Sigma' \subseteq \mathcal{I}(C,T)$. Hence $\mathcal{I}(C,T) \models D\sigma$. □

Lemma 3 cannot be generalized to the case where $C$ contains function symbols of arity $\geq 1$, take $C = P(f(x),y) \leftarrow P(z,x)$ and $D = P(f(a),a) \leftarrow P(a,f(a))$ (from the example given on p. 25 of Idestam-Almquist, 1993). Then $T = \{a, f(a)\}$ is the term set of $D$ and we have $C \models D$, yet it can be seen that $\mathcal{I}(C,T) \not\models D$. The argument used in the previous lemma does not work here, because different terms in some ground instance need not relate to different variables. For example, in the ground instance $P(f^2(a),a) \leftarrow P(a,f(a))$ of $C$, we cannot just replace $f^2(a)$ by some other term, for then the resulting clause would not be an instance of $C$.

On the other hand, Lemma 3 can be generalized to a *set* of clauses instead of a single clause. If $\Sigma$ is a set of function-free clauses, $C$ is an arbitrary clause, and $\sigma$ is a Skolem substitution for $C$ w.r.t. $\Sigma$, then we have that $\Sigma \models C$ iff $\mathcal{I}(\Sigma,T) \models C\sigma$. The proof is almost literally the same as above.

This result implies that $\Sigma \models C$ is reducible to an implication $\mathcal{I}(\Sigma,T) \models C\sigma$ between ground clauses. Since, by the next lemma, implication between ground clauses is decidable, it follows that $\Sigma \models C$ is decidable in case $\Sigma$ is function-free.

**Lemma 4** *The problem whether $\Sigma \models C$, where $\Sigma$ is a finite set of ground clauses and $C$ is a ground clause, is decidable.*

**Proof** Let $C = L_1 \vee \ldots \vee L_n$ and $\mathcal{A}$ be the set of all ground atoms occurring in $\Sigma$ and $C$. Now consider the following statements, which can be shown equivalent.
(1) $\Sigma \models C$.
(2) $\Sigma \cup \{\neg L_1, \ldots, \neg L_n\}$ is unsatisfiable.
(3) $\Sigma \cup \{\neg L_1, \ldots, \neg L_n\}$ has no Herbrand model.
(4) No subset of $\mathcal{A}$ is an Herbrand model of $\Sigma \cup \{\neg L_1, \ldots, \neg L_n\}$.

Then (1)$\Leftrightarrow$(2). (2)$\Leftrightarrow$(3) by Theorem 4.2 of (Chang & Lee, 1973). Since also (3)$\Leftrightarrow$(4), we have (1)$\Leftrightarrow$(4). (4) is decidable because $\mathcal{A}$ is finite, so (1) is decidable as well. □

**Corollary 1** *The problem whether $\Sigma \models C$, where $\Sigma$ is a finite set of function-free clauses and $C$ is a clause, is decidable.*

The following sequence of lemmas more or less follows the pattern of Idestam-Almquist's (1995) Lemma 10 to Lemma 12 (similar to Lemma 3.10 to Lemma 3.12 of Idestam-Almquist, 1993). There he gives a proof of the existence of a least generalization under T-implication of any finite set of (not necessarily function-free) clauses. We can adjust the proof in such a way that we can use it to establish the existence of an LGI of any finite set of clauses containing at least one non-tautologous function-free clause.

**Lemma 5** *Let $S$ be a finite set of non-tautologous clauses, $V = \{x_1, \ldots, x_m\}$ be a set of variables and let $G = \{C_1, C_2, \ldots\}$ be a (possibly infinite) set of generalizations of $S$ under implication. Then the set $G' = \mathcal{I}(C_1, V) \cup \mathcal{I}(C_2, V) \cup \ldots$ is a finite set of clauses.*

**Proof** Let $d$ be the maximal depth of the terms in clauses in $S$. It follows from Lemma 1 that $G$ (and hence also $G'$) cannot contain terms of depth greater than $d$, nor predicates, functions or constants other than those in $S$. The set of literals which can be constructed from predicates in $S$ and from terms of depth at most $d$ consisting of functions and constants in $S$ and variables in $V$, is finite. Hence the set of clauses which can be constructed from those literals is finite as well. $G'$ is a subset of this set, so $G'$ is a finite set of clauses. □

**Lemma 6** *Let $D$ be a clause, $C$ be a function-free clause such that $C \models D$, $T = \{t_1, \ldots, t_n\}$ be the term set of $D$ by $\sigma$, $V = \{x_1, \ldots, x_m\}$ be a set of variables and $m \geq n$. If $E$ is an LGS of $\mathcal{I}(C, V)$, then $E \models D$.*

**Proof** Let $\gamma = \{x_1/t_1, \ldots, x_n/t_n, x_{n+1}/t_n, \ldots, x_m/t_n\}$ (it does not matter to which terms the variables $x_{n+1}, \ldots, x_m$ are mapped by $\gamma$, as long as they are mapped to terms in $T$). Suppose $\mathcal{I}(C, V) = \{C\rho_1, \ldots, C\rho_k\}$. Then $\mathcal{I}(C, T) = \{C\rho_1\gamma, \ldots, C\rho_k\gamma\}$. Let $E$ be an LGS of $\mathcal{I}(C, V)$ (note that $E$ must be function-free). Then for every $1 \leq i \leq k$, there are $\theta_i$ such that $E\theta_i \subseteq C\rho_i$. This means that $E\theta_i\gamma \subseteq C\rho_i\gamma$ and hence $E\theta_i\gamma \models C\rho_i\gamma$, for every $1 \leq i \leq k$. Therefore $E \models \mathcal{I}(C, T)$.

Since $C \models D$, we know from Lemma 1 that constants appearing in $C$ must also appear in $D$. This means that $\sigma$ is a Skolem substitution for $D$ w.r.t. $\{C\}$. Then from Lemma 3 we know $\mathcal{I}(C, T) \models D\sigma$, hence $E \models D\sigma$. Furthermore, since $E$ is an LGS of $\mathcal{I}(C, V)$, all constants in $E$ also appear in $C$, hence all constants in $E$ must appear in $D$. Thus $\sigma$ is also a Skolem substitution for $D$ w.r.t. $\{E\}$. Therefore $E \models D$ by Lemma 2. □

Consider $C = P(x, y, z) \leftarrow P(y, z, x)$ and $D = \leftarrow Q(w)$. Both $C$ and $D$ imply the clause $E = P(x, y, z) \leftarrow P(z, x, y), Q(b)$. Now note that $C \cup D = P(x, y, z) \leftarrow P(y, z, x), Q(w)$ also implies $E$. This holds for clauses in general, even in the presence of background knowledge $\Sigma$. The next lemma is very general, but in this section we only need the special case where $C$ and $D$ are function-free and $\Sigma$ is empty. We need the general case to prove the existence of a GSR in Section 8.

**Lemma 7** *Let $C$, $D$ and $E$ be clauses such that $C$ and $D$ are standardized apart and let $\Sigma$ be a set of clauses. If $C \models_\Sigma E$ and $D \models_\Sigma E$, then $C \cup D \models_\Sigma E$.*

**Proof** Suppose $C \models_\Sigma E$ and $D \models_\Sigma E$, and let $M$ be a model of $\Sigma \cup \{C \cup D\}$. Since $C$ and $D$ are standardized apart, the clause $C \cup D$ is equivalent to the formula $\forall(C) \vee \forall(D)$ (where $\forall(C)$ denotes the universally quantified clause $C$). This means that $M$ is a model of $C$ or a model of $D$. Furthermore, $M$ is also a model of $\Sigma$, so it follows from $\Sigma \cup \{C\} \models E$ or $\Sigma \cup \{D\} \models E$ that $M$ is a model of $E$. Thus $\Sigma \cup \{C \cup D\} \models E$, hence $C \cup D \models_\Sigma E$. □

Now we can prove the existence of an LGI of any finite set $S$ of clauses which contains at least one non-tautologous and function-free clause. In fact we can prove something stronger, namely that this LGI is a *special* LGI. This is an LGI that is not only implied, but actually *subsumed* by any other generalization of $S$:

**Definition 17** Let $\mathcal{C}$ be a clausal language and $S$ be a finite subset of $\mathcal{C}$. An LGI $C$ of $S$ in $\mathcal{C}$ is called a *special* LGI of $S$ in $\mathcal{C}$, if $C' \succeq C$ for every generalization $C' \in \mathcal{C}$ of $S$ under implication. □

Note that if $D$ is an LGI of a set containing at least one non-tautologous function-free clause, then by Lemma 1 $D$ is itself function-free, because it should imply the function-free clause(s) in $S$. For instance, $C = P(x,y,z) \leftarrow P(y,z,x), Q(w)$ is an LGI of $D_1 = P(x,y,z) \leftarrow P(y,z,x), Q(f(a))$ and $D_2 = P(x,y,z) \leftarrow P(z,x,y), Q(b)$. Note that this LGI is properly subsumed by the LGS of $\{D_1, D_2\}$, which is $P(x,y,z) \leftarrow P(x',y',z'), Q(w)$. An LGI may sometimes be the empty clause □, for example if $S = \{P(a), Q(a)\}$.

**Theorem 6 (Existence of special LGI in $\mathcal{C}$)** *Let $\mathcal{C}$ be a clausal language. If $S$ is a finite set of clauses from $\mathcal{C}$ and $S$ contains at least one non-tautologous function-free clause, then there exists a special LGI of $S$ in $\mathcal{C}$.*

**Proof** Let $S = \{D_1, \ldots, D_n\}$ be a finite set of clauses from $\mathcal{C}$, such that $S$ contains at least one non-tautologous function-free clause. We can assume without loss of generality that $S$ contains no tautologies. Let $\sigma$ be a Skolem substitution for $S$, $T = \{t_1, \ldots, t_m\}$ be the term set of $S$ by $\sigma$, $V = \{x_1, \ldots, x_m\}$ be a set of variables and $G = \{C_1, C_2, \ldots\}$ be the set of all generalizations of $S$ under implication in $\mathcal{C}$. Note that $\Box \in G$, so $G$ is not empty. Since each clause in $G$ must imply the function-free clause(s) in $S$, it follows from Lemma 1 that all members of $G$ are function-free. By Lemma 5, the set $G' = \mathcal{I}(C_1, V) \cup \mathcal{I}(C_2, V) \cup \ldots$ is a finite set of clauses. Since $G'$ is finite, the set of $\mathcal{I}(C_i, V)$s is also finite. For simplicity, let $\{\mathcal{I}(C_1, V), \ldots, \mathcal{I}(C_k, V)\}$ be the set of all distinct $\mathcal{I}(C_i, V)$s.

Let $E_i$ be an LGS of $\mathcal{I}(C_i, V)$, for every $1 \leq i \leq k$, such that $E_1, \ldots, E_k$ are standardized apart. For every $1 \leq j \leq n$, the term set of $D_j$ by $\sigma$ is some set $\{t_{j_1}, \ldots, t_{j_s}\} \subseteq T$, such that $m \geq j_s$. From Lemma 6, we have that $E_i \models D_j$, for every $1 \leq i \leq k$ and $1 \leq j \leq n$, hence $E_i \models S$. Now let $F = E_1 \cup \ldots \cup E_k$, then we have $F \models S$ from Lemma 7 (applying the case of Lemma 7 where $\Sigma$ is empty).

To prove that $F$ is a special LGI of $S$, it remains to show that $C_j \succeq F$, for every $j \geq 1$. For every $j \geq 1$, there is an $i$ ($1 \leq i \leq k$), such that $\mathcal{I}(C_j, V) = \mathcal{I}(C_i, V)$. So for this $i$, $E_i$ is an LGS of $\mathcal{I}(C_j, V)$. $C_j$ is itself also a generalization of $\mathcal{I}(C_j, V)$ under subsumption, hence $C_j \succeq E_i$. Then finally $C_j \succeq F$, since $E_i \subseteq F$. □

As a consequence, we also immediately have the following:

**Corollary 2 (Existence of LGI for function-free clauses)** *Let $\mathcal{C}$ be a clausal language. Then for every finite set of function-free clauses $S \subseteq \mathcal{C}$, there exists an LGI of $S$ in $\mathcal{C}$.*

**Proof** Let $S$ be a finite set of function-free clauses in $\mathcal{C}$. If $S$ only contains tautologies, any tautology will be an LGI of $S$. Otherwise, let $S'$ be obtained by deleting all tautologies from $S$. By the previous theorem, there is a special LGI of $S'$. Clearly, this is also a special LGI of $S$ itself in $\mathcal{C}$. □

This corollary is not trivial, since even though the number of Herbrand interpretations of a language without function symbols is finite (due to the fact that the number of all possible ground atoms is finite in this case), $S$ may nevertheless be implied by an infinite number of non-equivalent clauses. This may seem like a paradox, since there are only finitely many categories of clauses that can "behave differently" in a *finite* number of finite Herbrand interpretations. Thus it would seem that the number of non-equivalent function-free clauses should also be finite. This is a misunderstanding, since logical implication (and hence also logical equivalence) is defined in terms of *all* interpretations, not just Herbrand interpretations. For instance, define $D_1 = P(a, a)$ and $P(b, b)$, $C_n = \{P(x_i, x_j) \mid i \neq j, 1 \leq i, j \leq n\}$. Then we have $C_n \models \{D_1, D_2\}$, $C_n \models C_{n+1}$ and $C_{n+1} \not\models C_n$, for every $n \geq 1$, see (van der Laag & Nienhuys-Cheng, 1994).

Another interesting consequence of Theorem 6 concerns self-saturation (see the introduction to this paper for the definition of self-saturation). If $C$ is a special LGI of some set $S$, then it is clear that $C$ is self-saturated: any clause which implies $C$ also implies $S$ and hence must *subsume* $C$, since $C$ is a special LGI of $S$. Now consider $S = \{D\}$, where $D$ is some non-tautologous function-free clause. Then a special LGI $C$ of $S$ will be logically equivalent to $D$. Moreover, since this $C$ will be self-saturated, it is a self-saturation of $D$.

**Corollary 3** *If $D$ is a non-tautologous function-free clause, then there exists a self-saturation of $D$.*

### 5.2 The LGI is Computable

In the previous subsection we proved the *existence* of an LGI in $\mathcal{C}$ of every finite set $S$ of clauses containing at least one non-tautologous function-free clause. In this subsection we will establish the *computability* of such an LGI. The next algorithm, extracted from the proof of the previous section, computes this LGI of $S$.

**LGI-Algorithm**

**Input:** A finite set $S$ of clauses, containing at least one non-tautologous function-free clause.
**Output:** An LGI of $S$ in $\mathcal{C}$.

1. Remove all tautologies from $S$ (a clause is a tautology iff it contains literals $A$ and $\neg A$), call the remaining set $S'$.
2. Let $m$ be the number of distinct terms (including subterms) in $S'$, let $V = \{x_1, \ldots, x_m\}$. (Notice that this $m$ is the same number as the number of terms in the term set $T$ used in the proof of Theorem 6.)
3. Let $G$ be the (finite) set of all clauses which can be constructed from predicates and constants in $S'$ and variables in $V$.
4. Let $\{U_1, \ldots, U_n\}$ be the set of all subsets of $G$.
5. Let $H_i$ be an LGS of $U_i$, for every $1 \leq i \leq n$. These $H_i$ can be computed by Plotkin's (1970) algorithm.
6. Remove from $\{H_1, \ldots, H_n\}$ all clauses which do not imply $S'$ (since each $H_i$ is function-free, by Corollary 1 this implication is decidable), and standardize the remaining clauses $\{H_1, \ldots, H_q\}$ apart.
7. Return the clause $H = H_1 \cup \ldots \cup H_q$.

The correctness of this algorithm follows from the proof of Theorem 6. First notice that $H \models S$ by Lemma 7. Furthermore, note that all $\mathcal{I}(C_i, V)$'s mentioned in the proof of Theorem 6, are elements of the set $\{U_1, \ldots, U_n\}$. This means that for every $E_i$ in the set $\{E_1, \ldots, E_k\}$ mentioned in that proof, there is a clause $H_j$ in $\{H_1, \ldots, H_q\}$ such that $E_i$ and $H_j$ are subsume-equivalent. Then it follows that the LGI $F = E_1 \cup \ldots \cup E_k$ of that proof subsumes the clause $H = H_1 \cup \ldots \cup H_q$ that our algorithm returns. On the other hand, $F$ is a special LGI, so $F$ and $H$ must be subsume-equivalent.

Suppose the number of distinct constants in $S'$ is $c$ and the number of distinct variables in step 2 of the algorithm is $m$. Furthermore, suppose there are $p$ distinct predicate symbols in $S'$, with respective arities $a_1, \ldots, a_p$. Then the number of distinct atoms that can be formed from these constants, variables and predicates, is $l = \sum_{i=1}^{p}(c+m)^{a_i}$, and the number of distinct literals that can be formed is $2 \cdot l$. The set $G$ of distinct clauses which can be formed from these literals is the power set of this set of literals, so $|G| = 2^{2 \cdot l}$. Then the set $\{U_1, \ldots, U_n\}$ of all subsets of $G$ contains $2^{|G|} = 2^{2^{2 \cdot l}}$ members.

Thus the algorithm outlined above is not very efficient (to say the least). A more efficient algorithm may exist, but since implication is harder than subsumption and the computation of an LGS is already quite expensive, we should not put our hopes too high. Nevertheless, the existence of the LGI-algorithm does establish the theoretical point that the LGI of any finite set of clauses containing at least one non-tautologous function-free clause is effectively computable.

**Theorem 7 (Computability of LGI)** *Let $\mathcal{C}$ be a clausal language. If $S$ is a finite set of clauses from $\mathcal{C}$, and $S$ contains at least one non-tautologous function-free clause, then the LGI of $S$ in $\mathcal{C}$ is computable.*

## 6. Greatest Specializations under Implication

Now we turn from least generalizations under implication to greatest specializations. Finding least generalizations of sets of clauses is common practice in ILP. On the other hand, the greatest specialization, which is the dual of the least generalization, is used hardly ever. Nevertheless, the GSI of two clauses $D_1$ and $D_2$ might be useful. Suppose that we have one positive example $e^+$ and two negative examples $e_1^-$ and $e_2^-$ and suppose that $D_1$ implies $e^+$ and $e_1^-$, while $D_2$ implies $e^+$ and $e_2^-$. Then it might very well be that the GSI of $D_1$ and $D_2$ still implies $e^+$, but does not imply either $e_1^-$ or $e_2^-$. Thus we could obtain a correct specialization by taking the GSI of $D_1$ and $D_2$.

It is obvious from the previous sections that the existence of an LGI of $S$ is quite hard to establish. For clauses which all contain functions, the existence of an LGI is still an open question, and even for the case where $S$ contains at least one non-tautologous function-free clause, the proof was far from trivial. However, the existence of a GSI in $\mathcal{C}$ is much easier to prove. In fact, a GSI of a finite set $S$ is the same as the GSS of $S$, namely the union of the clauses in $S$ after these are standardized apart.

To see the reason for this dissymmetry, let us take a step back from the clausal framework and consider full first-order logic for a moment. If $\phi_1$ and $\phi_2$ are two arbitrary first-order formulas, then it can be easily shown that their least generalization is just $\phi_1 \wedge \phi_2$: this conjunction implies $\phi_1$ and $\phi_1$, and must be implied by any other formula which implies both $\phi_1$ and $\phi_2$. Dually, the greatest specialization is just $\phi_1 \vee \phi_2$: this is implied by both $\phi_1$ and $\phi_2$, and must imply any other formula that is implied by both $\phi_1$ and $\phi_2$. See figure 3.

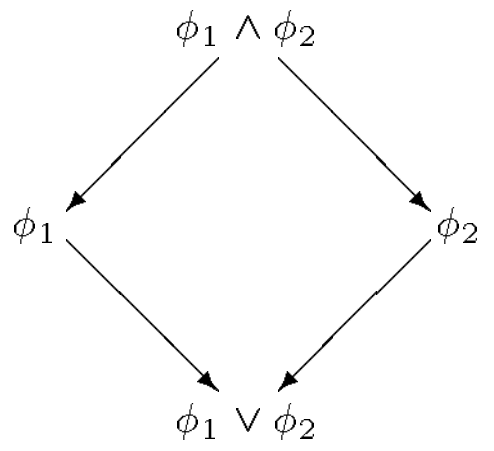

Figure 3: Least generalization and greatest specialization in first-order logic

Now suppose $\phi_1$ and $\phi_2$ are clauses. Then why do we have a problem in finding the LGI of $\phi_1$ and $\phi_2$? The reason for this is that $\phi_1 \wedge \phi_2$ is not a clause. Instead of using $\phi_1 \wedge \phi_2$, we have to find some least clause which implies both clauses $\phi_1$ and $\phi_2$. Such a clause appears quite hard to find sometimes.

On the other hand, in case of specialization there is no problem. Here we can take $\phi_1 \vee \phi_2$ as GSI, since $\phi_1 \vee \phi_2$ is equivalent to a clause, if we handle the universal quantifiers in front of a clause properly. If $\phi_1$ and $\phi_2$ are standardized apart, then the formula $\phi_1 \vee \phi_2$ is equivalent to the clause which is the union of $\phi_1$ and $\phi_2$. This fact was used in the proof of Lemma 7.

Suppose $S = \{D_1, \ldots, D_n\}$, and $D_1', \ldots, D_n'$ are variants of these clauses which are standardized apart. Then clearly $D = D_1' \cup \ldots \cup D_n'$ is a GSI of $S$, since it follows from Lemma 7 that any specialization of $S$ under implication is implied by $D$. Thus we have the following result:

**Theorem 8 (Existence of GSI in $\mathcal{C}$)** *Let $\mathcal{C}$ be a clausal language. Then for every finite $S \subseteq \mathcal{C}$, there exists a GSI of $S$ in $\mathcal{C}$.*

The previous theorem holds for clauses in general, so in particular also for function-free clauses. Furthermore, Corollary 2 guarantees us that in a function-free clausal language an LGI of every finite $S$ exists. This means that the set of function-free clauses quasi-ordered by logical implication is in fact a lattice.

**Corollary 4 (Lattice-structure of function-free clauses under $\models$)** *A function-free clausal language ordered by implication is a lattice.*

In case of a Horn language $\mathcal{H}$, we cannot apply the same proof method as in the case of a clausal language, since the union of two Horn clauses need not be a Horn clause itself. In fact, we can show that not every finite set of Horn clauses has a GSI in $\mathcal{H}$. Here we can use the same clauses that we used to show that sets of Horn clauses need not have an LGI in $\mathcal{H}$, this time from the perspective of specialization instead of generalization.

Again, let $D_1 = P(f^2(x)) \leftarrow P(x)$, $D_2 = P(f^3(x)) \leftarrow P(x)$, $C_1 = P(f(x)) \leftarrow P(x)$ and $C_2 = P(f^2(y)) \leftarrow P(x)$. Then $C_1 \models \{D_1, D_2\}$ and $C_2 \models \{D_1, D_2\}$, and there is no Horn clause $D$ such that $D \models D_1$, $D \models D_2$, $C_1 \models D$ and $C_2 \models D$. Hence there is no GSI of $\{C_1, C_2\}$ in $\mathcal{H}$.

## 7. Least Generalizations under Relative Implication

Implication is stronger than subsumption, but *relative* implication is even more powerful, because background knowledge can be used to model all sorts of useful properties and relations. In this section, we will discuss least generalizations under implication relative to some given background knowledge $\Sigma$ (LGR's). In the next section we treat greatest specializations under relative implication.

First, we will prove the equivalence between our definition of relative implication and a definition given by Niblett (1988, p. 133). He gives the following definition of subsumption relative to a background knowledge $\Sigma$ (to distinguish it from our notion of subsumption, we will call this 'N-subsumption'):[9]

**Definition 18** Clause $C$ *N-subsumes* clause $D$ with respect to background knowledge $\Sigma$ if there is a substitution $\theta$ such that $\Sigma \vdash (C\theta \rightarrow D)$ (here '$\rightarrow$' is the implication-connective, and '$\vdash$' is an arbitrary complete proof procedure). □

**Proposition 1** *Let $C$ and $D$ be clauses and $\Sigma$ be a set of clauses. Then $C$ N-subsumes $D$ with respect to $\Sigma$ iff $C \models_\Sigma D$.*

**Proof** Consider the following six statements, which can be shown equivalent.
(1) $C$ N-subsumes $D$ with respect to $\Sigma$.
(2) There is a substitution $\theta$ such that $\Sigma \vdash (C\theta \rightarrow D)$.
(3) There is a substitution $\theta$ such that $\Sigma \models (C\theta \rightarrow D)$.

9. Niblett attributes this definition to Plotkin, though Plotkin gives a rather different definition of relative subsumption in (Plotkin, 1971b), as we have seen in Section 4.

(4) There is a substitution $\theta$ such that $\Sigma \cup \{C\theta\} \models D$.
(5) $\Sigma \cup \{C\} \models D$.
(6) $C \models_\Sigma D$.
(1)$\Leftrightarrow$(2) by definition. (2)$\Leftrightarrow$(3) by the completeness of $\vdash$. (3)$\Leftrightarrow$(4) by the Deduction Theorem. (4)$\Rightarrow$(5) is obvious and (5)$\Rightarrow$(4) follows from letting $\theta$ be the empty substitution, hence (4)$\Leftrightarrow$(5). Finally, (5)$\Leftrightarrow$(6) by definition. Thus these six statements are equivalent. □

Since $\models$ is the special case of $\models_\Sigma$ where $\Sigma$ is empty, our counterexamples to the existence of LGI's or GSI's in $\mathcal{H}$ are also counterexamples to the existence of LGR's or GSR's in $\mathcal{H}$. In other words, the '$-$'-entries in the second row of Table 1 carry over to the third row.

For general clauses, the LGR-question also has a negative answer. We will show here that even if $S$ and $\Sigma$ are both finite sets of *function-free* clauses, an LGR of $S$ relative to $\Sigma$ need not exist. Let $D_1 = P(a)$, $D_2 = P(b)$, $S = \{D_1, D_2\}$, and $\Sigma = \{(P(a) \vee \neg Q(x)), (P(b) \vee \neg Q(x))\}$. We will show that this $S$ has no LGR relative to $\Sigma$ in $\mathcal{C}$.

Suppose $C$ is an LGR of $S$ relative to $\Sigma$. Note that if $C$ contains the literal $P(a)$, then the Herbrand interpretation which makes $P(a)$ true and which makes all other ground literals false, would be a model of $\Sigma \cup \{C\}$ but not of $D_2$, so then we would have $C \not\models_\Sigma D_2$. Similarly, if $C$ contains $P(b)$ then $C \not\models_\Sigma D_1$. Hence $C$ cannot contain $P(a)$ or $P(b)$.

Now let $d$ be a constant not appearing in $C$. Let $D = P(x) \vee Q(d)$, then $D \models_\Sigma S$. By the definition of an LGR, we should have $D \models_\Sigma C$. Then by the subsumption theorem, there must be a derivation from $\Sigma \cup \{D\}$ of a clause $E$, which subsumes $C$. The set of all clauses which can be derived (in 0 or more resolution-steps) from $\Sigma \cup \{D\}$ is $\Sigma \cup \{D\} \cup \{(P(a) \vee P(x)), (P(b) \vee P(x))\}$. But none of these clauses subsumes $C$, because $C$ does not contain the constant $d$, nor the literals $P(a)$ or $P(b)$. Hence $D \not\models_\Sigma C$, contradicting the assumption that $C$ is an LGR of $S$ relative to $\Sigma$ in $\mathcal{C}$.

Thus in general the LGR of $S$ relative to $\Sigma$ need not exist. However, we can identify a special case in which the LGR *does* exist. This case might be of practical interest. Suppose $\Sigma = \{L_1, \ldots, L_m\}$ is a finite set of *function-free ground* literals. We can assume $\Sigma$ does not contain complementary literals (i.e., $A$ and $\neg A$), for otherwise $\Sigma$ would be inconsistent. Also, suppose $S = \{D_1, \ldots, D_n\}$ is a set of clauses, at least one of which is non-tautologous and function-free. Then $C \models_\Sigma D_i$ iff $\{C\} \cup \Sigma \models D_i$ iff $C \models D_i \vee \neg(L_1 \wedge \ldots \wedge L_m)$ iff $C \models D_i \vee \neg L_1 \vee \ldots \vee \neg L_m$. This means that an LGI of the set of clauses $\{(D_1 \vee \neg L_1 \vee \ldots \vee \neg L_m), \ldots, (D_n \vee \neg L_1 \vee \ldots \vee \neg L_m)\}$ is also an LGR of $S$ relative to $\Sigma$. If some $D_k \vee \neg L_1 \vee \ldots \vee \neg L_m$ is non-tautologous and function-free, this LGI exists and is computable. Hence in this special case, the LGR of $S$ relative to $\Sigma$ exists and is computable.

## 8. Greatest Specializations under Relative Implication

Since the counterexample to the existence of GSI's in $\mathcal{H}$ is also a counterexample to the existence of GSR's in $\mathcal{H}$, the only remaining question in the $\models_\Sigma$-order is the existence of GSR's in $\mathcal{C}$. The answer to this question is positive. In fact, like the GSS and the GSI, the GSR of some finite set $S$ in $\mathcal{C}$ is just the union of the (standardized apart) clauses in $S$.

**Theorem 9 (Existence of GSR in $\mathcal{C}$)** *Let $\mathcal{C}$ be a clausal language and $\Sigma \subseteq \mathcal{C}$. Then for every finite $S \subseteq \mathcal{C}$, there exists a GSR of $S$ relative to $\Sigma$ in $\mathcal{C}$.*

**Proof** Suppose $S = \{D_1, \ldots, D_n\} \subseteq \mathcal{C}$. Without loss of generality, we assume the clauses in $S$ are standardized apart. Let $D = D_1 \cup \ldots \cup D_n$, then $D_i \models_\Sigma D$, for every $1 \leq i \leq n$. Now let $C \in \mathcal{C}$ be such that $D_i \models_\Sigma C$, for every $1 \leq i \leq n$. Then from Lemma 7, we have $D \models_\Sigma C$. Hence $D$ is a GSR of $S$ relative to $\Sigma$ in $\mathcal{C}$. □

## 9. Conclusion

In ILP, the three main generality orders are subsumption, implication, and relative implication. The two main languages are clausal languages and Horn languages. This gives a total of six different ordered sets. In this paper, we have given a systematic treatment of the existence or non-existence of least generalizations and greatest specializations in each of these six ordered sets. The outcome of this investigation is summarized in Table 1. The only remaining open question is the existence or non-existence of a least generalization under implication in $\mathcal{C}$ for sets of clauses which all contain function symbols.

Table 1 makes explicit the trade-off between different generality orders. On the one hand, implication is better suited as a generality order than subsumption, particularly in case of recursive clauses. Relative implication is still better, because it allows us to take background knowledge into account. On the other hand, we can see from the table that as far as the existence of least generalizations goes, subsumption is more attractive than logical implication, and logical implication is in turn more attractive than relative implication. For subsumption, least generalizations always exist. For logical implication, we can only prove the existence of least generalizations in the presence of a function-free clause. And finally, for relative implication, least generalizations need not even exist in a function-free language. In practice this means that we cannot have it all. If we choose to use a very strong generality order such as relative implication, least generalizations only exist in very limited cases. On the other hand, if we want to guarantee that least generalizations always exist, we are committed to the weakest generality order: subsumption.

## Acknowledgements

We would like to thank Peter Idestam-Almquist and the referees for their comments, which helped to improve the paper.